\documentclass[10pt,twocolumn,letterpaper]{article}

\usepackage[pagenumbers]{iccv} 
\usepackage[accsupp]{axessibility}
\usepackage{tabularx}
\usepackage{xcolor}
\usepackage{pifont}
\newcommand{\cmark}{\ding{51}}%
\newcommand{\xmark}{\ding{55}}%
\definecolor{iccvblue}{rgb}{0.21,0.49,0.74}
\usepackage[pagebackref,breaklinks,colorlinks,allcolors=iccvblue]{hyperref}

\def\paperID{*****} 
\def\confName{ICCV}
\def\confYear{2025}

\title{VQualA 2025 Challenge on Visual Quality Comparison for Large Multimodal Models: Methods and Results}

\author{Hanwei Zhu$^{\dagger}$ \and Haoning Wu$^{\dagger}$ \and Zicheng Zhang$^{\dagger}$ \and Lingyu Zhu$^{\dagger}$ \and Yixuan Li$^{\dagger}$ \and Peilin Chen$^{\dagger}$ \and Shiqi Wang$^{\dagger}$ \and Chris Wei Zhou$^{\dagger}$ \and Linhan Cao \and Wei Sun \and Xiangyang Zhu \and Weixia Zhang \and Yucheng Zhu \and Jing Liu \and Dandan Zhu \and Guangtao Zhai \and Xiongkuo Min \and Zhichao Zhang \and Xinyue Li \and Shubo Xu \and Anh Dao \and Yifan Li \and Hongyuan Yu \and Jiaojiao Yi \and Yiding Tian \and Yupeng Wu \and Feiran Sun \and Lijuan Liao \and Song Jiang
}

\begin{document}
\maketitle

\renewcommand{\thefootnote}{}
\footnotetext{$^{\dagger}$Hanwei Zhu, Haoning Wu, Zicheng Zhang, Lingyu Zhu, Yixuan Li, Peilin Chen, Shiqi Wang, and Chris Wei Zhou are the challenge organizers.  
(Corresponding authors: \textit{Shiqi Wang} (shiqwang@cityu.edu.hk)).}
\footnotetext{The other authors are participants of the VQualA 2025 Challenge on Visual Quality Comparison for Large Multimodal Models.}
\footnotetext{The VQualA 2025 website:~\url{https://vquala.github.io//}}
\footnotetext{The Competition website~\url{https://codalab.lisn.upsaclay.fr/competitions/23016}}
\footnotetext{The Co-Instruct-562K dataset:~\url{https://co-instruct.github.io/}}

\begin{abstract}
This paper presents a summary of the VQualA 2025 Challenge on Visual Quality Comparison for Large Multimodal Models (LMMs), hosted as part of the ICCV 2025 Workshop on Visual Quality Assessment. The challenge aims to evaluate and enhance the ability of state-of-the-art LMMs to perform open-ended and detailed reasoning about visual quality differences across multiple images. To this end, the competition introduces a novel benchmark comprising thousands of coarse-to-fine grained visual quality comparison tasks, spanning single images, pairs, and multi-image groups. Each task requires models to provide accurate quality judgments. The competition emphasizes holistic evaluation protocols, including 2AFC-based binary preference and multi-choice questions (MCQs). Around 100 participants submitted entries, with five models demonstrating the emerging capabilities of instruction-tuned LMMs on quality assessment. This challenge marks a significant step toward open-domain visual quality reasoning and comparison and serves as a catalyst for future research on interpretable and human-aligned quality evaluation systems. 

\end{abstract}

\section{Introduction}
\label{sec:intro}
Image quality assessment (IQA) plays a fundamental role in low-level vision and multimedia applications~\cite{chen2022beyond,chen2023compact,hui2023rate,hui2024s}, enabling the development of systems that can evaluate and enhance the perceptual quality of visual content~\cite{10886996,fang2020perceptual,lin2011perceptual,dists}. Traditional IQA methods predominantly regress scalar scores, such as mean opinion scores (MOS), to assess individual images. However, these absolute quality ratings often suffer from limited interpretability and inconsistency, especially in subjective or fine-grained scenarios where human perception varies across observers~\cite{Wu_2024_CVPR,wu2024towards,zhu2afc24}. In contrast, comparative judgments, asking which image among a group is sharper, more natural, or less distorted, have been shown to yield more consistent and informative assessments in both psychophysical studies and real-world applications~\cite{wu2024towards,zhu2afc24,kong2025pixel,kong2022detect}.


With the emergence of Large Multimodal Models (LMMs) such as GPT-4o~\cite{hurst2024gpt}, Claude-3.5-Sonnet~\cite{anthropic2024claude35}, LLaVA-OneVision-7B~\cite{li2024llava}, and Qwen2.5-VL-7B~\cite{bai2025qwen2},
there has been increasing interest in leveraging these powerful reasoning engines for perceptual tasks. Recent works demonstrate that LMMs exhibit promising capabilities in visual quality understanding when prompted appropriately, particularly in two-alternative forced choice (2AFC) settings~\cite{zhu2afc24,wu2024comprehensive}. Nevertheless, current benchmarks and datasets primarily target single-image analysis or pairwise judgments~\cite{tian2025ai}, lacking the complexity and flexibility required for real-world comparative IQA.

To advance this frontier, we present the Visual Quality Comparison track of the VQualA 2025 Challenge, hosted at the first VQualA workshop in conjunction with ICCV 2025. This challenge aims to systematically evaluate and push the limits of LMMs in open-ended, multi-image quality comparison tasks. Participants are required to build models that can reason over multiple images—pairs, triplets, or quadruplets—and generate both quality decisions and natural language explanations. For example, given a set of distorted images, models must answer nuanced queries such as “Which image is overexposed?” or “Rank these four images in terms of sharpness, and explain why.” 

The competition is powered by Co-Instruct-562K~\cite{wu2024towards}, the first large-scale instruction-tuning dataset designed for visual quality comparison. Constructed via a hybrid strategy of weak supervision, Co-Instruct-562K combines (1) Merge2Compare, which synthesizes comparisons from human-annotated single-image descriptions~\cite{Wu_2024_CVPR}, and (2) Teach2Compare, which leverages GPT-4V to generate pseudo-labels on unlabeled multi-image groups. To enable rigorous evaluation, we also introduce multiple images comparison benchmark~(MICBench), a new benchmark consisting of 4,000 human-annotated multiple-choice questions (MCQs) spanning groups of two, three, or four images and covering diverse distortion types and quality attributes. In particular, we construct the pairwise comparisons following the coarse-to-fine pairing protocol proposed in~\cite{zhu2afc24}, which ensures that image pairs exhibit either distinct distortion levels or types while maintaining comparable content to facilitate fine-grained perceptual judgment. For groups of three or four images, we extend this principle by inheriting the structural design and task format of MICBench to preserve inter-image comparability while enhancing the challenge and diversity of the visual quality discrimination task~\cite{wu2024towards}. These group-level questions are formulated using semantic-aware rules and validated through controlled human annotation procedures to ensure label reliability. Collectively, the benchmark enables systematic investigation into LMMs’ ability to generalize across comparison cardinalities and distortion modalities, while providing a robust foundation for training and evaluation of instruction-tuned quality assessment models.

The Visual Quality Comparison track of the VQualA 2025 Challenge attracted a total of 92 registered participants. Throughout the development phase, 373 submissions were received, followed by 538 prediction results during the final testing phase. In the end, 5 valid teams submitted final models and fact sheets for this track. Each participating team provided a detailed description of its LMM-based visual quality comparison methods. A comprehensive summary of the challenge results is presented in Sec. \ref{sec:cha_res} and \ref{sec:teams}.

This track aims not only to benchmark the current capabilities of LMMs in comparative IQA but also to inspire future research at the intersection of vision, language, and perception. By integrating structured evaluation protocols with open-ended reasoning tasks, the Visual Quality Comparison track establishes a new direction for interpretable, human-aligned quality assessment in multi-image and video scenarios. This challenge is one of the VQualA 2025 Workshop associated challenges on: face image quality assessment~\cite{ma2025fiqa}, image super-resolution generated content quality assessment~\cite{isrgcq2025iccvw}, engagement prediction for short videos~\cite{li2025evqa}, document image enhancement quality assessment~\cite{diqa2025iccvw}, GenAI-bench AIGC video quality assessment track I~\cite{genai-bench2025iccvw}, and GenAI-Bench AIGC video quality assessment track II~\cite{genai-bench2025iccvw}. In the following sections, we describe the challenge in detail, present and analyze the results, and provide an overview of the participating methods.

\section{Related Work}
\label{sec:rw}
\subsection{Traditional IQA}
Image quality assessment (IQA) is critically important in numerous multimedia and visual communication applications, as visual content frequently undergoes various forms of distortion during processes such as capturing~\cite{zhu2022learning,fang2017perceptual,sui2023perceptual,kong2024moe}, compression~\cite{zhu2024video,chen2025pleno}, transmission~\cite{liu2023quality,yan2022subjective}, and restoration~\cite{fang2021superpixel,zhu2024unrolled, zhu2024temporally,chen2023gap,zhao2018faster}. Early objective metrics treated IQA as a signal-fidelity problem. Peak-signal-to-noise ratio (PSNR)~\cite{psnr} establishes a numeric baseline but correlates weakly with human perception. The seminal Structural-Similarity Index (SSIM)~\cite{ssim,wang2004image} reframed fidelity as comparisons of luminance, contrast, and structural information, spawning multi-scale and information-weighted variants that remain influential today. In parallel, the Visual-Information Fidelity (VIF)~\cite{sheikh2006image} criterion modeled quality as mutual information preserved by a distortion channel, linking IQA to information theory and further improving correlation with MOSs. Deep learning then supplanted manual features, such as the learned perceptual image patch similarity (LPIPS)~\cite{LPIPS18}, deep image structure and texture similarity~(DISTS)~\cite{dists}, the deep Wasserstein distance~(DeepWSD)~\cite{deepwsd}, the deep Distance correlation (DeepDC)~\cite{zhu2022deepdc}. Moreover, when it comes to the no-reference IQA, CNNIQA~\cite{kang2014convolutional} demonstrated that a shallow patch-based convolutional network can regress MOS directly from raw pixels. Deeper or more adaptive variants followed, including bilinear pooling in DBCNN~\cite{dbcnn}, content-conditioned HyperIQA~\cite{hyperiqa}, and generalized statistics method~\cite{chen2024deep}, while transformers such as MUSIQ~\cite{ke2021musiq} and Swin-IQA~\cite{wang2022mstriq} provided stronger global reasoning for native-resolution inputs. Self-supervised paradigms such as CONTRIQUE~\cite{madhusudana2022image} further reduced dependence on labeled data by learning quality-aware representations through contrastive objectives and auxiliary distortion prediction.

Since absolute MOS labels are costly, rank-based learning emerged as an efficient alternative~\cite{liqe,unique}. RankIQA~\cite{liu2017rankiqa} trained a Siamese network on synthetically distorted pairs and then transferred the learned embedding for single-image scoring, outperforming regression-only counterparts when annotations are scarce. The PaQ-2-PiQ~\cite{ying2020patches} dataset pushed this idea further by collecting large-scale human pairwise preferences enabling deep ranking losses that map distance in feature space to perceptual order and yielding state-of-the-art global and local quality estimates. Further, researchers began to study perceptual quality comparison itself rather than scalar scoring. PieAPP~\cite{prashnani2018pieapp} learns a perceptual error function directly from human pairwise preferences and predicts the probability that one image is preferred over another. Such development shifts the focus from absolute fidelity to relative and contextual judgment, laying the conceptual groundwork for the multimodal, reasoning-driven quality-comparison approaches focused on this challenge.
\subsection{LMM-based IQA}
The integration of LMMs into IQA has recently gained traction, particularly due to their proficiency in capturing semantic context through multi-modal representation learning. Early efforts to transfer LMMs to IQA started from prompt- or adapter-based CLIP adaptations. Zhang \textit{et al.}~\cite{liqe} showed that a naive prompt such as ``a good photo'' lets CLIP offer surprisingly strong zero-shot correlations, while the performance is highly prompt-sensitive. Subsequent work therefore learned soft prompts or lightweight adapters, where the recent GRMP-IQA framework~\cite{li2024boosting} went a step further by pre-training visual–text meta-prompts and regularizing gradients so that the visual branch focuses on distortion instead of visual semantics, which achieved strong few-shot generalization capability. These studies establish that the rich semantic priors of vision–language models can be unlocked for quality perception without task-specific heads.

Supervised fine-tuning (SFT) then became mainstream. Q-Instruct~\cite{Wu_2024_CVPR} constructed a 200K quality-related visual instruction corpus converted from human low-level feedback and instruction-tuned LLaVA-style bases. The model can answer fine-grained questions and regress an overall score in a single forward pass. Later, Q-Align~\cite{qalign} argued that humans learn discrete text-defined rating levels and therefore align LMMs to five textual levels, which unifies IQA, image-aesthetic assessment, and video QA in one model. The proposed model yielded higher quality prediction performance than the direct quality score regression. Regarding the open-ended image quality comparison task, Co-Instruct~\cite{wu2024towards} contributed the Merge2Compare and Teach2Compare quality comparison corpus and a fine-tuned visual-language model that answers free-form comparative questions with explicit rationales. On the other hand, Compare2Score~\cite{zhu2024adaptive} taught LMMs to perform pairwise quality comparisons synthesized from MOS ground-truth, and inference the probability that a test image beats several anchors to derive a continuous quality score via MAP estimation. It bridges comparative quality reasoning with single-image regression. Parallel to these, DepictQA~\cite{you2024depicting} shifted the target from numbers to descriptive sentences, enabling more adaptive open-ended quality diagnosis. Later, its successor DeQA-Score~\cite{you2025teaching} converted those natural-language judgments into a score-distribution supervision that lets LLMs regress accurate MOS while remaining explainable. 

Most recently, reinforcement-learning (RL) variants have been explored to endow LMMs with stronger visual reasoning abilities. Benefited from RL, Q-Insight~\cite{qinsight} formulated IQA as a two-headed task containing quality score regression and degradation perception, and optimized both with group relative policy optimization (GRPO). The carefully designed rewards let the model learn with limited labels yet generalize to zero-shot quality comparison tasks. VisualQuality-R1~\cite{wu2025visualquality} further exploited GRPO but casted IQA as ranking, where for each image pair, multiple quality estimates are sampled, mapped to Thurstone comparative probabilities, and rewarded with continuous fidelity signals. The resulting model not only topped traditional NR-IQA models but also wrote human-aligned rationales. To encapsulate, the trajectory of LMM-based IQA has progressed from prompt-tuned CLIP baselines to instruction-tuned assessors, distribution- and comparison-aware regressors, and finally reasoning-induced RL agents. This field is moving from score prediction to quality understanding, where numeric fidelity, relative ranking, and human-readable explanations are jointly optimized within a single foundation model.
\section{Challenge Configurations}
\label{sec:cha_data}
In this section, we first introduce the Co-Instruct-562K training corpus in detail. Subsequently, we describe the evaluation protocols, including the validation and testing sets of MICBench and the metric used in the challenge.

\begin{figure*}[t]
    \centering
    \subfloat[AWGN\_Level-$1$]{\includegraphics[width=0.156\linewidth]{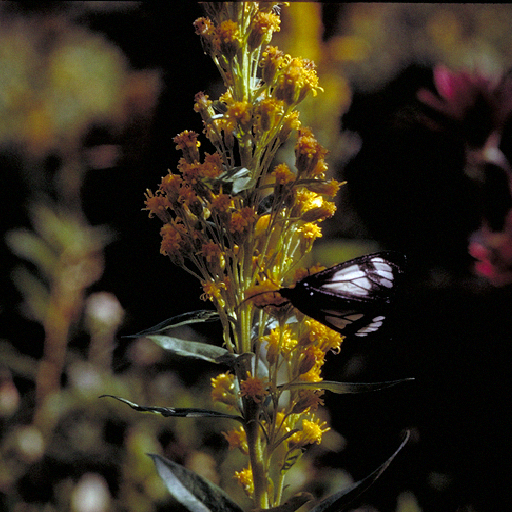}}\hskip.1em
    \subfloat[AWGN\_Level-$2$]{\includegraphics[width=0.156\linewidth]{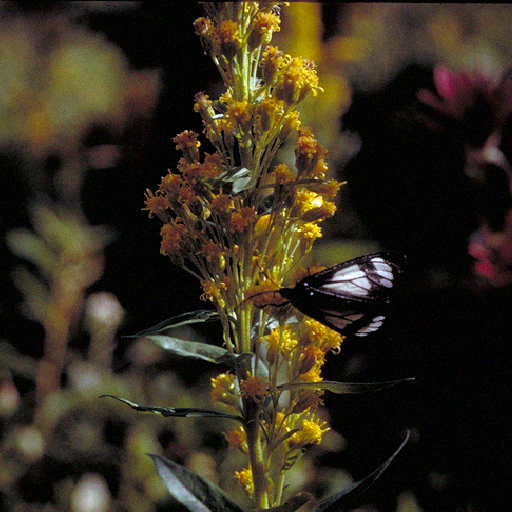}}\hskip.1em
    \subfloat[JP2K\_Level-$4$]{\includegraphics[width=0.156\linewidth]{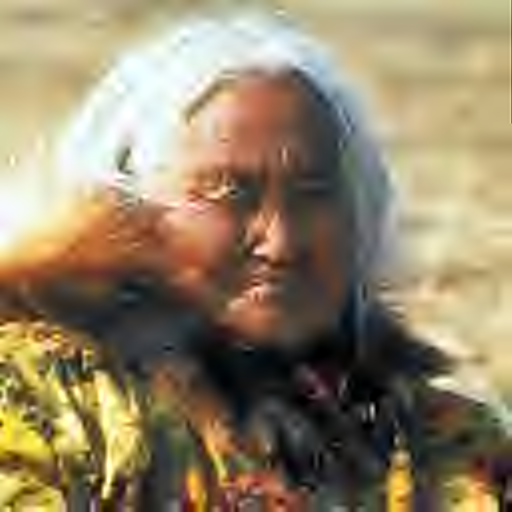}}\hskip.1em
    \subfloat[Pink\_Level-$4$]{\includegraphics[width=0.156\linewidth]{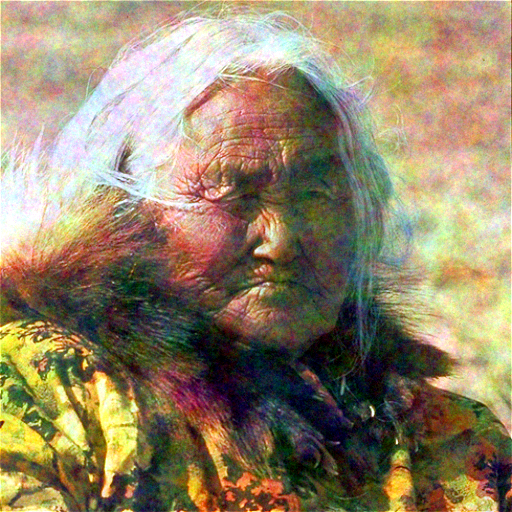}}\hskip.1em
    \subfloat[MOS = $14$]{\includegraphics[width=0.173\linewidth]{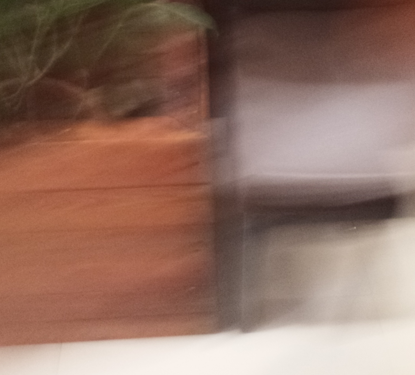}}\hskip.1em
    \subfloat[MOS = $23$]{\includegraphics[width=0.173\linewidth]{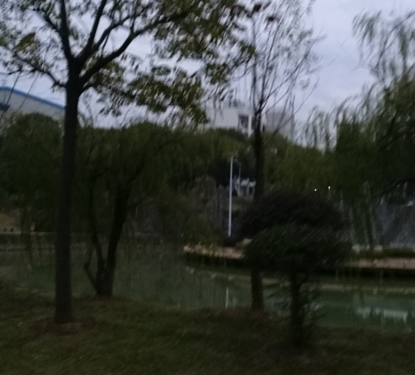}}
    \caption{Illustration of three pairing rules for fine-grained quality comparison. \textbf{(a)}\&\textbf{(b)} Two synthetically distorted images with identical visual content and distortion type but different distortion levels. \textbf{(c)}\&\textbf{(d)} Two synthetically distorted images with identical visual content and distortion level but different distortion types. \textbf{(e)}\&\textbf{(f)} Two realistically distorted images in the MOS interval of $[0, 25)$.  Image by courtesy of~\cite{zhu2afc24}. 
 }
    \label{fig:fine}
\end{figure*}

\begin{figure*}[h]
    \centering
    \includegraphics[width=0.9\linewidth]{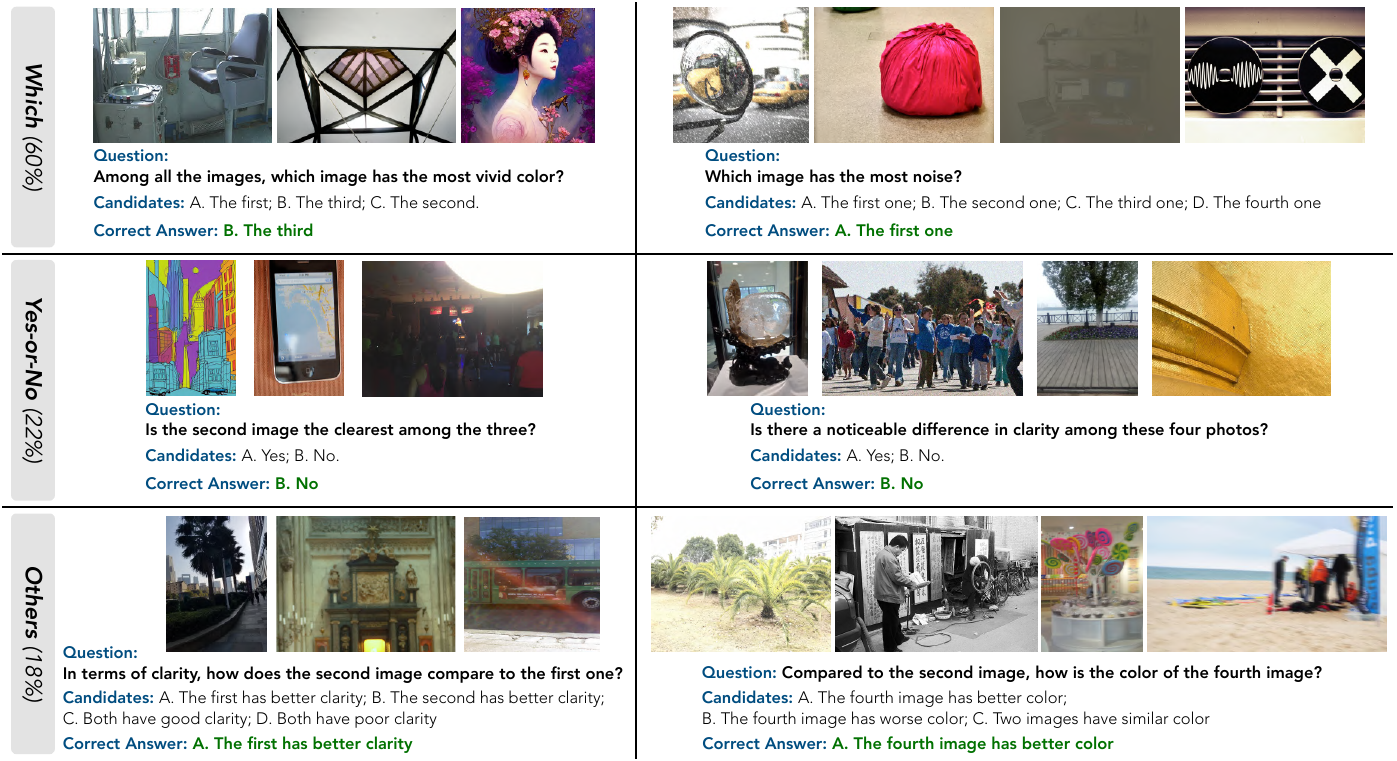}
    \caption{Dataset Card of MIC-Bench, made up of \textbf{(a)} \textit{Which} questions, \textbf{(b)} \textit{Yes-or-No} questions, and \textbf{(c) }\textit{Other} types of questions on three/four images. Image by courtesy of~\cite{wu2024towards}. }
    \label{fig:micmcq}
\end{figure*}

\subsection{Co-Instruct-562K}

The training data used in the Visual Quality Comparison track of the VQualA 2025 Challenge is Co-Instruct-562K, the first large-scale instruction-tuning dataset explicitly designed for open-ended visual quality comparison. This dataset addresses the limitations of traditional scalar-based IQA annotations by providing rich, natural language supervision across various image grouping levels—enabling both fine-grained visual discrimination and linguistic reasoning.

Co-Instruct-562K consists of four major subsets:
\begin{itemize}
\item \textbf{202K single-image instructions:} Derived from Q-Instruct-200K, this subset provides high-quality textual descriptions of individual images across diverse distortion types and quality attributes (e.g., blur, compression, overexposure).
\item \textbf{222K image pairs:} Each pair is accompanied by comparative questions and explanations, such as “Which image is sharper?” or “Which one has fewer artifacts, and why?” These comparisons allow models to learn relative quality assessment beyond absolute scoring.
\item \textbf{77K triplets:} Groups of three images are annotated with ranking or selection tasks, enabling models to learn listwise comparative reasoning across multiple distorted instances.
\item \textbf{61K quadruplets:} These involve four-image comparisons, focusing on nuanced ranking, identification of worst/best cases, or structured reasoning across complex quality differences.
\end{itemize}

The dataset construction follows a hybrid pipeline combining human-generated and model-generated supervision:
\begin{itemize}
\item \textbf{Merge2Compare:} This process merges independently written human single-image descriptions into coherent group-wise comparative instructions using GPT-based prompting. It ensures logical consistency and contextual relevance across comparisons.
\item \textbf{Teach2Compare:} To enrich coverage, GPT-4V is employed as a teacher model to generate pseudo-comparative annotations on unlabeled image groups. These include natural language justifications and question-answer formats designed to mirror human-like comparative reasoning.
\end{itemize}

Overall, Co-Instruct-562K represents a comprehensive, instruction-tuned resource that supports the training of LMMs capable of performing interpretable, multi-image quality comparisons. It serves as the foundation for enabling open-domain IQA capabilities with grounding in both visual and linguistic modalities.

\subsection{Evaluation Protocols}

The evaluation set for the Visual Quality Comparison track is MICBench, a curated benchmark specifically designed to assess the fine-grained comparative reasoning abilities of LMMs. While Co-Instruct-562K enables open-ended instruction-tuning during training, MICBench provides a structured, standardized evaluation framework through multiple-choice questions (MCQs) constructed over multi-image groups.

MICBench consists of two main subsets:
\begin{itemize}
\item \textbf{Validation Set~\footnote{\url{https://huggingface.co/datasets/VQA-CityU/MICBench-val}}:} 2,000 MCQs released for model validation, consisting of 1004 image pairs, 494 image triplets, and 502 image quadruplets.
\item \textbf{Test Set~\footnote{\url{https://huggingface.co/datasets/VQA-CityU/MICBench-test}}:} 2,000 MCQs used for final evaluation in the VQualA 2025 Challenge leaderboard, consisting of 1016 image pairs, 504 image triplets, and 480 image quadruplets.
\end{itemize}

\begin{table*}[t]
\centering
\renewcommand\arraystretch{1.2}
\tabcolsep=0.3cm
\small
\begin{tabular}{c|c|c|cccc|c}
\toprule
\textbf{Rank} & \textbf{Team} & \textbf{Leader} & \textbf{Final} & \textbf{\# Two} & \textbf{\# Three} & \textbf{\# Four} & \textbf{Extra Data} \\
\midrule
1 & ECNU-SJTU VQA Team & Linhan Cao & 0.7570 & 0.6713 & 0.8452 & 0.8458 & \xmark\\
2 & Digital Ocean & Zhichao Zhang & 0.7560 & 0.6550 & 0.8710 & 0.8479 & \xmark \\
3 & ActionLab & Anh Dao & 0.7555 & 0.6841 & 0.8234 & 0.8345 & \xmark\\
4 & XiaomiMM & Hongyuan Yu & 0.7510 & 0.6860 & 0.8353 & 0.8000 & \xmark \\
5 & Labubu & Lijuan Jiao & 0.7255 & 0.6693 & 0.7798 & 0.7875 & \cmark\\
\hline
\multicolumn{3}{c|}{LLaVA-OneVision-7B} & 0.7335 & 0.6575 & 0.8214 & 0.8021 & N/A \\
\multicolumn{3}{c|}{Qwen2.5-VL-7B} & 0.7340 & 0.6526 & 0.8115 & 0.8250 & N/A \\
\multicolumn{3}{c|}{Qwen2.5-VL-72B} & 0.7550 & 0.6654 & 0.8472 & 0.8479 & N/A \\
\bottomrule
\end{tabular}
\caption{Leaderboard results including team rankings and baseline performances.}
\label{tab:leaderboard}
\end{table*}

Each multiple-choice question (MCQ) in MICBench presents a comparative query grounded in groups of 2, 3, or 4 images, encompassing a broad spectrum of distortion types such as Gaussian noise, blur, compression artifacts, color shifts, and overexposure. To rigorously evaluate the IQA capabilities of large multimodal models (LMMs), we introduce a set of systematic coarse-to-fine pairing strategies, as illustrated in Fig.~\ref{fig:fine}. For coarse-grained quality comparison, images are randomly paired from the same dataset, ensuring variability in both content and distortion type. In contrast, for fine-grained quality comparison, we propose three structured pairing rules:
\begin{enumerate}
\item \textbf{Same content and distortion type, different distortion levels:} Synthetic distortions are applied at varying severity levels to identical visual content, enabling models to assess subtle quality degradations.
\item \textbf{Same content and distortion level, different distortion types:} Images are paired such that the visual content and distortion intensity remain fixed, while the distortion type varies—encouraging models to reason about distortion-specific perceptual impacts.
\item \textbf{Realistic distortions within the same MOS interval:} Images affected by real-world distortions are selected such that their subjective quality scores fall within the same MOS bin, presenting nuanced perceptual comparisons of similarly rated content.
\end{enumerate}

For image groups of three and four, we inherit the MCQ construction protocol from MICBench in Co-Instruct~\cite{wu2024towards}. As depicted in Fig.~\ref{fig:micmcq}, the questions are carefully formulated to elicit high-level perceptual reasoning and multi-step natural language understanding. The distribution of question types includes:
\begin{itemize}
\item \textbf{“Which” questions:} e.g., “Which image has the most noticeable motion blur?” or “Which of the four images is sharpest?”
\item \textbf{Yes/No questions:} e.g., “Is the second image oversaturated?” or “Does image A exhibit banding artifacts?”
\item \textbf{What/How/Why questions:} e.g., “How many images exhibit color inconsistency?” or “What type of distortion affects the third image?”
\end{itemize}

All MCQs are authored and verified by expert annotators to ensure perceptual validity, linguistic clarity, and consistent task difficulty. The answer formats are intentionally diverse, ranging from categorical selections to binary responses to assess the general reasoning capabilities of LMMs across task formats.


Evaluation is conducted using top-1 accuracy across all MCQs. Submissions are scored by comparing the model-predicted choice against the ground-truth answer key. This protocol facilitates objective model comparison while minimizing the ambiguity often present in open-ended textual evaluation. In addition, the structured design of MICBench enables granular analysis by distortion type, image group size, and question format, offering deeper insights into the reasoning capabilities and perceptual fidelity of participating models.

\section{Challenge Results}
\label{sec:cha_res}

The Visual Quality Comparison track of the VQualA 2025 Challenge garnered significant attention from the research community, attracting a total of 92 registered participants. During the competition phases, participants actively engaged in developing and fine-tuning their models, resulting in 373 submissions in the development phase and 538 predictions during the final testing phase. Ultimately, five teams successfully submitted their final models along with comprehensive fact sheets, showcasing detailed methodologies leveraging state-of-the-art LMM for fine-grained visual quality assessment.

Table~\ref{tab:leaderboard} summarizes the final leaderboard, highlighting the top-performing teams and baseline methods across different configurations of image comparisons—pairs, triplets, and quadruplets. Team \textbf{ECNU-SJTU VQA} secured first place with an overall accuracy of 0.7570. Their ensemble-based solution, named FGVQA, demonstrated robust and consistent performance across two-image (0.6713), three-image (0.8452), and four-image (0.8458) comparative tasks, underscoring the effectiveness of combining multiple LMMs through majority voting. The \textbf{Digital Ocean} team closely followed in second place, achieving an overall accuracy of 0.7560. Their innovative approach utilized joint optimization via supervised fine-tuning (SFT) and direct preference optimization (DPO), enhanced by an ensemble voting strategy. This resulted in notably high accuracies on three-image (0.8710) and four-image (0.8479) comparisons, reflecting strong multi-modal reasoning capabilities. \textbf{ActionLab} secured third place with a commendable accuracy of 0.7555. They employed a two-stage fine-tuning methodology on the Qwen2.5-VL-7B model, achieving balanced performance across different image group sizes. The fourth-place team, \textbf{XiaomiMM} introduced a Multi-stage Enhanced Visual Reasoning Architecture (MEVRA), effectively integrating auxiliary quality scores and structured prompt engineering, achieving an accuracy of 0.7510. Team \textbf{Labubu} completed the top five rankings with an accuracy of 0.7255. Their approach emphasized quality-aware embedding enhancements and explicit ranking constraints, leading to substantial improvements in fine-grained perceptual discrimination.

Additionally, baseline methods including LLaVA-OneVision-7B (0.7335), Qwen2.5-VL-7B (0.7340), and the more robust Qwen2.5-VL-72B (0.7550) provided strong reference points, validating the advanced performance of the top teams. These results illustrate substantial advancements in visual quality comparison powered by multimodal reasoning and instruction-tuned models, highlighting the potential for further innovation in this promising research direction.

\section{Teams and Methods}
\label{sec:teams}

\subsection{ECNU-SJTU VQA}
\begin{figure}[!t]
	\centering
	\includegraphics[width=0.48\textwidth]{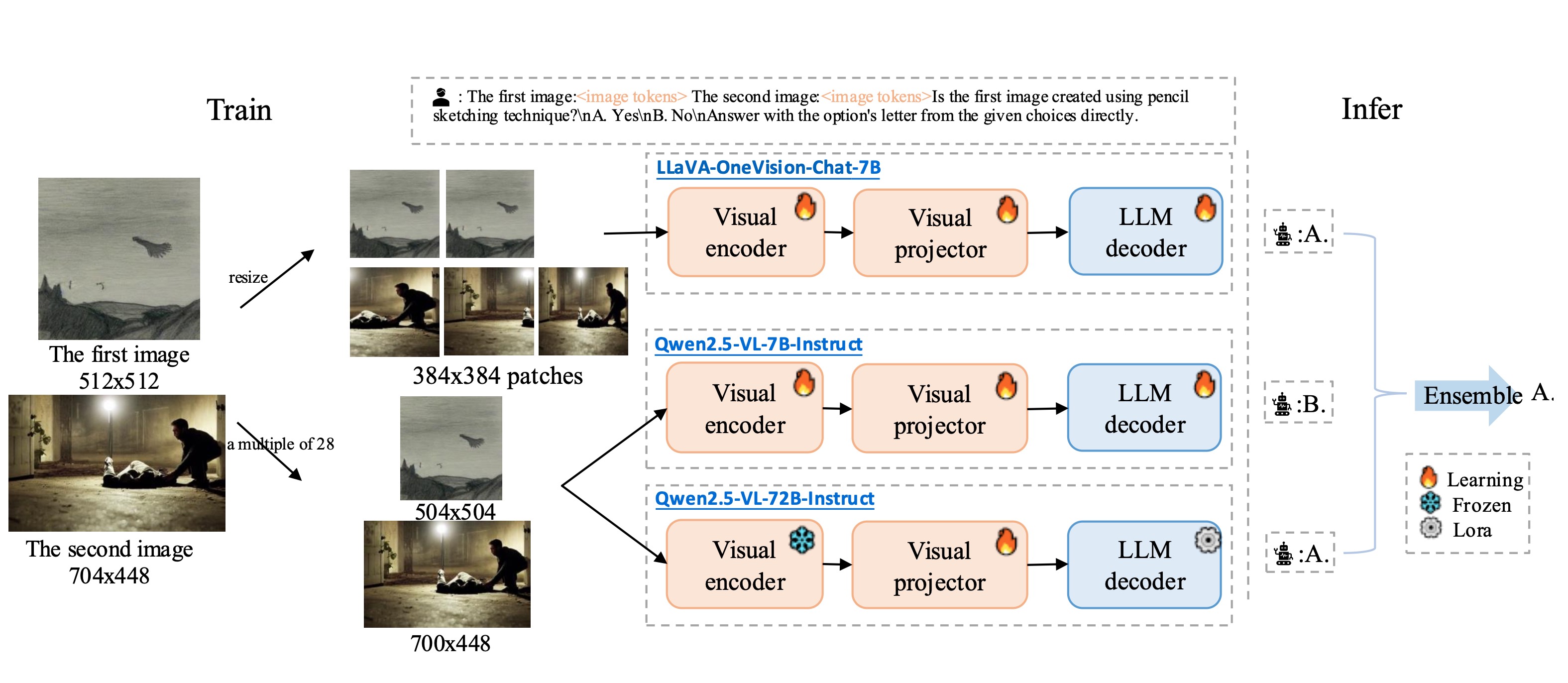}
	\caption{The overall framework of Team ECNU-SJTU VQA.}
\label{fig:overview_1}
\end{figure}
This team develops an ensemble-based framework, FGVQA, for fine-grained visual quality comparison. Recognizing the limitations of scalar image quality scores and the subjectivity of traditional IQA models, the team integrates three LMMs, including LLaVA-OneVision-Chat-7B, Qwen2.5-VL-7B-Instruct, and Qwen2.5-VL-72B-Instruct, to enhance reasoning-driven and open-ended visual quality assessment. Each model is independently prompted on the same image group, and their responses are aggregated using a consensus-driven majority voting strategy. This ensemble approach exploits complementary model capacities, enabling scalable and human-aligned comparative IQA.

\noindent \textbf{Training Details:} 
All three LMMs are trained exclusively on the Co-Instruct-562K dataset using the AdamW optimizer, with no external data utilized. For LLaVA-OneVision-Chat-7B, each input image is resized to 384×384 and divided into patches for the vision encoder; the model is fine-tuned with all parameters updated, using a batch size of 8 on 8 A800 GPUs for one epoch (vision encoder LR: 2e-6, rest: 1e-5). For Qwen2.5-VL-7B-Instruct, image pixels are limited (max 1280×28×28) for efficient memory use, and all parameters are trained on 8 A800 GPUs with a batch size of 16 for one epoch (vision encoder LR: 2e-6, rest: 1e-5). For Qwen2.5-VL-72B-Instruct, only the visual projector and the LLM (via LoRA) are tuned, following the same image preprocessing as the 7B model, with LoRA rank and alpha both set to 64, dropout 0.05, vision LR 1e-5, and LoRA LR 1e-4.

\noindent \textbf{Testing Details:} 
For evaluation, each model uses the same image preprocessing as during training. The models independently predict their answers (\textit{e.g.,} A/B/C/D) for each question, and the final output is determined by majority voting. If all three models provide different answers, the result from Qwen2.5-VL-72B-Instruct is adopted, leveraging its superior reasoning ability. The ensemble is executed in parallel on 3 A800 GPUs: the two 7B models share one GPU, while the 72B model uses two GPUs. The primary metric is overall ensemble answer accuracy. Experiments show that the ensemble method achieves the highest accuracy, outperforming individual models and confirming the effectiveness of model fusion for fine-grained visual quality assessment.

\subsection{Digital Ocean}
\begin{figure}[!t]
	\centering
	\includegraphics[width=0.48\textwidth]{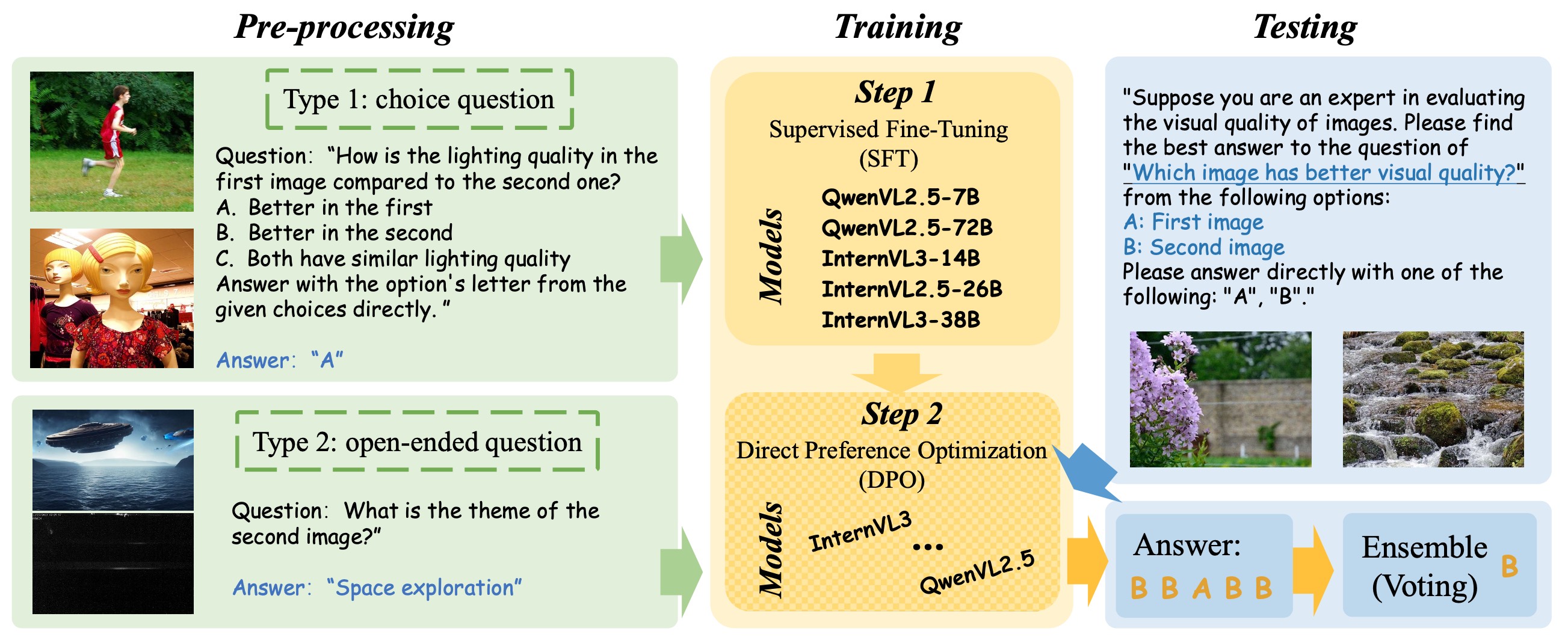}
	\caption{The overall framework of Team Digital Ocean.}
\label{fig:overview_2}
\end{figure}
This team proposes a joint optimization framework for visual quality comparison by leveraging an ensemble of five state-of-the-art MLLMs: InternVL3-14B, InternVL2.5-26B, InternVL3-38B, QwenVL2.5-7B, and QwenVL2.5-72B. The proposed approach incorporates both Supervised Fine-Tuning (SFT) and Direct Preference Optimization (DPO), enhancing the model’s ability to provide accurate and human-aligned answers to both multiple-choice and open-ended visual quality comparison questions. To further boost robustness and reduce model bias, a voting-based ensemble strategy is adopted, aggregating the predictions from all five fine-tuned models to generate the final output.

\noindent \textbf{Training Details:}
The team first screens and preprocesses a large-scale image-text dataset, filtering out samples with lengthy responses to ensure training effectiveness. In the SFT stage, the models are fine-tuned on supervised data, while the DPO stage further optimizes the models with preference data focused on multi-choice tasks. LoRA is applied to fine-tune only the language components, keeping the vision encoders frozen for computational efficiency. Training is performed on 4 H20 GPUs with a batch size of 1 per device, for 1 epoch and a learning rate of 1e-4, using bfloat16 precision and gradient accumulation. The LoRA rank is set to 8 and alpha to 32, with the MSE loss function, warmup ratio of 0.05, and a maximum input sequence length of 4096 tokens. No external data is used beyond the provided data.

\noindent \textbf{Testing Details:} 
During testing, each question is reformulated in a unified prompt template. The five trained MLLMs individually predict their answers to each question, and a voting-based method is used to determine the final ensemble output. If a tie occurs, preference is given to the outputs from QwenVL2.5-72B and InternVL3-38B. The framework achieves robustness and accuracy by integrating the strengths of all models and averaging their predictions for binary classification tasks. The approach is tested on an H20 GPU with a typical input size (1000, 1000, 3), and the model operates without any post-processing.

\subsection{ActionLab}
This team develops a solution for open-ended visual quality comparison, based on the two-stage fine-tuning of the Qwen2.5-VL-7B large multimodal model. Their approach leverages a frozen-vision warm-up phase followed by full-model fine-tuning, inspired by techniques used in recent multimodal LLMs such as LLaVA and MiniGPT-4. The method directly outputs the answer letter in open-ended comparison tasks without extra post-processing.

\noindent \textbf{Training Details:} The training process consists of two main stages. In the first stage (frozen-vision), only the language backbone and multimodal projector are updated, while the CLIP-ViT vision encoder is kept frozen. In the second stage, the entire model is fine-tuned end-to-end. The team used PyTorch 2.3, HuggingFace Transformers, and DeepSpeed ZeRO-3 for training, running on 8 NVIDIA A6000 GPUs (48GB each). Optimization is performed with AdamW (LR 2e-7 for the LLM, 1e-5 for the projector), a cosine learning rate schedule, and 3\% warm-up. The bfloat16 precision and gradient checkpointing are applied for memory and efficiency optimization, with FlashAttention-2 and fused Adam for further speedup. The team trained exclusively on the official Co-Instruct-562K dataset, avoiding any external IQA data. Stage 1 took 100 hours, and Stage 2 took 24 hours, totaling 124 hours of training.

\noindent \textbf{Testing Details:} Inference is performed on a single NVIDIA H100 GPU, achieving an average latency of just 0.28 seconds per MICBench question. Greedy decoding (temperature=0) is adopted, as experiments with self-consistency and temperature sampling yielded lower accuracy and higher latency. The model outputs predictions directly, with no ensemble applied.

\subsection{XiaomiMM}
\begin{figure}[!t]
	\centering
	\includegraphics[width=0.5\textwidth]{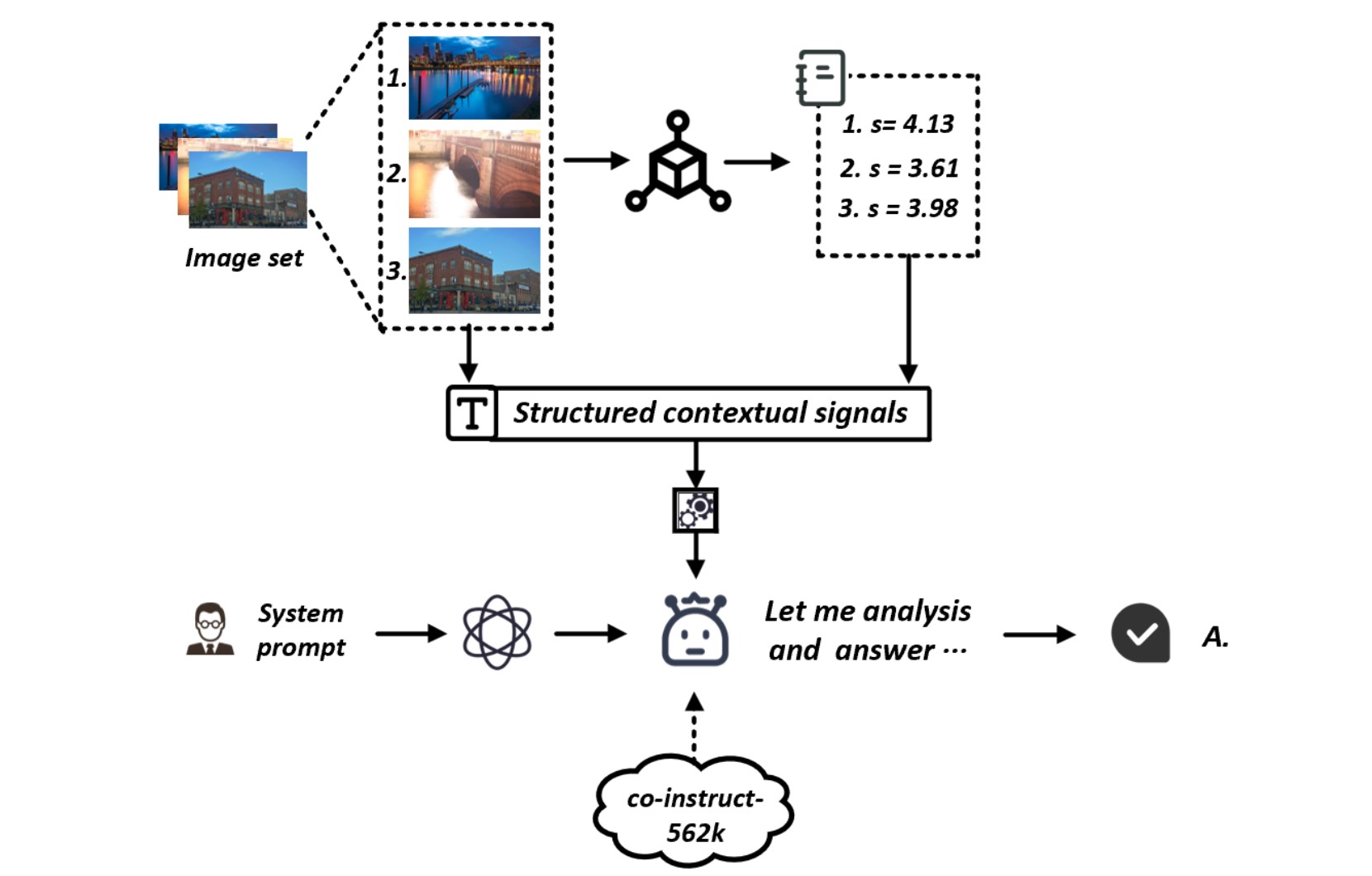}
	\caption{The overall framework of Team XiaomiMM.}
\label{fig:overview_3}
\end{figure}

This team develops a Multi-stage Enhanced Visual Reasoning Architecture (MEVRA) designed to boost the multi-image reasoning abilities of large multimodal models for visual quality comparison tasks. The framework is built upon the high-capacity Qwen2.5-VL-72B model and integrates auxiliary visual scoring to provide explicit quality-related signals, enhancing the model's ability to handle complex quality comparison scenarios involving multiple images. MEVRA operates in three stages: a coarse-grained fine-tuning stage to adapt the model for image quality analysis, an auxiliary scoring injection stage that introduces structured quality scores into the prompt, and a task-driven cognitive alignment stage that optimizes the model’s attention to relative image differences.

\noindent \textbf{Training Details:}
The XiaomiMM team employs LoRA-based fine-tuning on all linear layers of the Qwen2.5-VL-72B model, using the large-scale Co-instruct-562K dataset. The data is split into a training and validation set (99:1 ratio). During the initial coarse-grained fine-tuning stage, LoRA injects knowledge of image quality perception into the backbone, improving the model’s baseline performance. In the auxiliary scoring injection stage, the lightweight quality assessment model q-align-4bit generates independent quality scores for each input image, which are then batch-injected as structured signals alongside the original vision-language inputs. The final task-driven cognitive alignment stage uses progressive prompt engineering and context control to optimize attention mechanisms for comparative reasoning. Training concludes after 17,380 steps, and the merged LoRA and weights are used for inference.

\noindent \textbf{Testing Details:} 
During inference, each image is first processed by the Q-align-4bit model to obtain quality scores. These scores are injected into the prompt template, and the enhanced prompt is then input to the MEVRA model for reasoning and decision output. The context control mechanism dynamically masks irrelevant quality signals in non-quality judgment tasks, effectively denoising the input and improving robustness. The final accuracy on the test dataset demonstrates the effectiveness of multi-stage reasoning and explicit quality cue integration for fine-grained visual quality comparison.

\subsection{Labubu}
\begin{figure}[!t]
	\centering
	\includegraphics[width=0.48\textwidth]{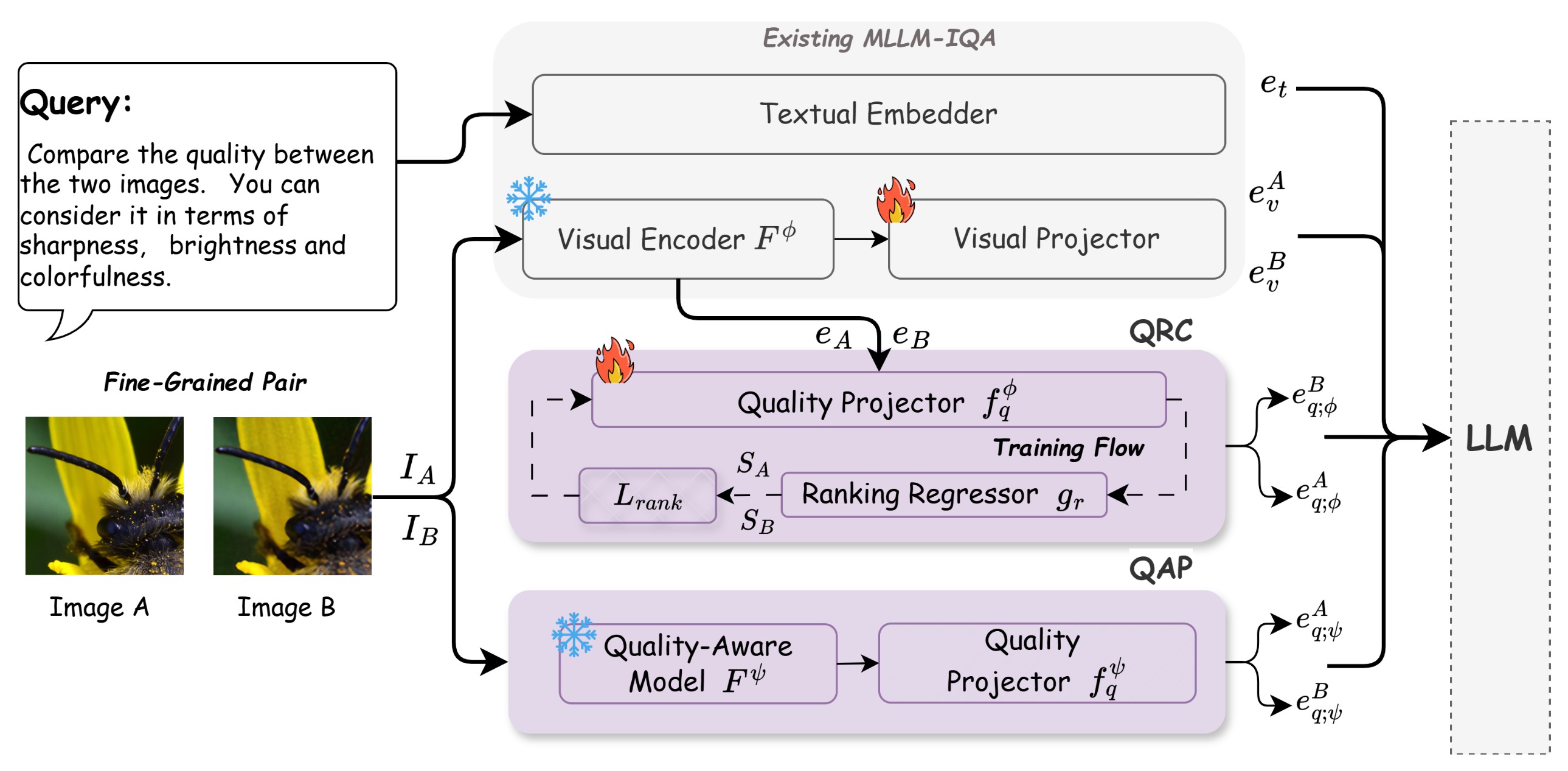}
	\caption{The overall framework of Team Labubu.}
\label{fig:overview_4}
\end{figure}

This team proposes a novel method for fine-grained visual quality comparison by enhancing MLLMs with explicit quality ranking and quality-aware features. Their approach, built upon DepictQA-Wild, tries to address the limitations of traditional autoregressive (AR) fine-tuning and semantic visual encoders by introducing two key modules: the Quality Ranking Constraint (QRC) and the Quality Aware Projection (QAP). The QRC module applies a margin ranking loss to amplify discernible differences in visual quality between image pairs, while the QAP module enriches the visual representation with features from pre-trained IQA models, such as ReIQA, CONTRIQUE, and ARNIQA. These quality-aware and ranking embeddings are fused with standard visual and textual embeddings, enabling the model to deliver improved fine-grained quality comparison.

\noindent \textbf{Training Details:} The proposed method extends DepictQA-Wild, leveraging a frozen CLIP-ViT-L/14 visual encoder and a Vicuna-7B large language model (LLM). The Quality Ranking Constraint (QRC) and Quality Aware Projection (QAP) modules are implemented as single-layer perceptrons, ensuring computational efficiency and lightweight integration. Training is performed jointly on these newly introduced modules and other trainable components, utilizing the Low-Rank Adaptation (LoRA) strategy consistent with DepictQA-Wild. Specifically, the QRC projector has an input dimension of 1,024, matching the CLIP output, while the QAP projector input dimension is set to 2,048 to accommodate quality-aware models. The main LoRA loss and ranking loss weights are empirically determined as 1.0 and 10.0, respectively. The entire training process is conducted over a single epoch on two NVIDIA A100-80G GPUs. The instruction fine-tuning dataset, Co-Instruct-562K, comprises four subsets: Q-Instruct-200K, \textit{Merge2Compare}, \textit{Teach2Compare}-general, and \textit{Teach2Compare}-QA. Specifically, the Q-Instruct-200K subset aligns with the single-image branch, with quality ranking labels designated as None. For the other subsets, answers from each sample are processed through the Ranking Label Generator to derive quality ranking labels in a structured Python list format, ordered by descending quality. However, for samples lacking clear ranking labels—such as those in the Teach2Compare-QA subset the quality ranking labels are likewise set to None.


\noindent \textbf{Testing Details:} The model is evaluated using the official challenge protocol, primarily measuring accuracy on MICBench-dev (multi-choice and open-ended questions). Inference follows the DepictQA-Wild baseline, with no additional post-processing. Results show that both QRC and QAP modules contribute to improved accuracy, and the use of additional quality-aware models and extra data yields further gains. 

\section*{Acknowledgements}
The research was partially supported by the RGC General Research Fund 11200323 and  NSFC/RGC JRS Project N\_CityU198/24.

{
    \small
    \bibliographystyle{ieeenat_fullname}
    \bibliography{main}

\begin{thebibliography}{63}
\providecommand{\natexlab}[1]{#1}
\providecommand{\url}[1]{\texttt{#1}}
\expandafter\ifx\csname urlstyle\endcsname\relax
  \providecommand{\doi}[1]{doi: #1}\else
  \providecommand{\doi}{doi: \begingroup \urlstyle{rm}\Url}\fi

\bibitem[{Anthropic}(2024)]{anthropic2024claude35}
{Anthropic}.
\newblock Claude 3.5 {S}onnet: Faster, smarter, and more useful.
\newblock \url{https://www.anthropic.com/news/claude-3-5-sonnet}, 2024.
\newblock Accessed: 2025-05-15.

\bibitem[Bai et~al.(2025)Bai, Chen, Liu, Wang, Ge, Song, Dang, Wang, Wang, Tang, et~al.]{bai2025qwen2}
Shuai Bai, Keqin Chen, Xuejing Liu, Jialin Wang, Wenbin Ge, Sibo Song, Kai Dang, Peng Wang, Shijie Wang, Jun Tang, et~al.
\newblock {Qwen2.5-VL} technical report.
\newblock \emph{arXiv preprint arXiv:2502.13923}, 2025.

\bibitem[Chen et~al.(2022)Chen, Wang, Li, Lin, Wang, and Ye]{chen2022beyond}
Bolin Chen, Zhao Wang, Binzhe Li, Rongqun Lin, Shiqi Wang, and Yan Ye.
\newblock Beyond keypoint coding: Temporal evolution inference with compact feature representation for talking face video compression.
\newblock In \emph{Data Compression Conference}, pages 13--22. IEEE, 2022.

\bibitem[Chen et~al.(2023{\natexlab{a}})Chen, Wang, Li, Wang, and Ye]{chen2023compact}
Bolin Chen, Zhao Wang, Binzhe Li, Shiqi Wang, and Yan Ye.
\newblock Compact temporal trajectory representation for talking face video compression.
\newblock \emph{IEEE Transactions on Circuits and Systems for Video Technology}, 33\penalty0 (11):\penalty0 7009--7023, 2023{\natexlab{a}}.

\bibitem[Chen et~al.(2023{\natexlab{b}})Chen, Zhu, Zhu, Yang, Song, and Wang]{chen2023gap}
Baoliang Chen, Lingyu Zhu, Hanwei Zhu, Wenhan Yang, Linqi Song, and Shiqi Wang.
\newblock Gap-closing matters: Perceptual quality evaluation and optimization of low-light image enhancement.
\newblock \emph{IEEE Transactions on Multimedia}, 26:\penalty0 3430--3443, 2023{\natexlab{b}}.

\bibitem[Chen et~al.(2024)Chen, Zhu, Zhu, Wang, and Kwong]{chen2024deep}
Baoliang Chen, Hanwei Zhu, Lingyu Zhu, Shiqi Wang, and Sam Kwong.
\newblock Deep feature statistics mapping for generalized screen content image quality assessment.
\newblock \emph{IEEE Transactions on Image Processing}, 2024.

\bibitem[Chen et~al.(2025{\natexlab{a}})Chen, Zhu, Yin, Zhu, Chen, Liao, Wang, and Ye]{chen2025pleno}
Bolin Chen, Hanwei Zhu, Shanzhi Yin, Lingyu Zhu, Jie Chen, Ru-Ling Liao, Shiqi Wang, and Yan Ye.
\newblock Pleno-generation: A scalable generative face video compression framework with bandwidth intelligence.
\newblock \emph{arXiv preprint arXiv:2502.17085}, 2025{\natexlab{a}}.

\bibitem[Chen et~al.(2025{\natexlab{b}})Chen, Zhu, Zhu, Wang, Pan, and Wang]{10886996}
Baoliang Chen, Hanwei Zhu, Lingyu Zhu, Shanshe Wang, Jingshan Pan, and Shiqi Wang.
\newblock Debiased mapping for full-reference image quality assessment.
\newblock \emph{IEEE Trans. Multimedia}, pages 1--12, 2025{\natexlab{b}}.

\bibitem[Chen et~al.(2025{\natexlab{c}})Chen, Wang, Xiao, Ding, Liu, Zhou, and et~al.]{genai-bench2025iccvw}
Ying Chen, Huasheng Wang, Pengxiang Xiao, Yukang Ding, Enpeng Liu, Chris~Wei Zhou, and et al.
\newblock Vquala 2025 challenge on genai-bench aigc video quality assessment: Methods and results.
\newblock In \emph{Proceedings of the IEEE/CVF Conference on Computer Vision (ICCV) Workshops}, pages 1--11, 2025{\natexlab{c}}.

\bibitem[Ding et~al.(2022)Ding, Ma, Wang, and Simoncelli]{dists}
Keyan Ding, Kede Ma, Shiqi Wang, and Eero~P. Simoncelli.
\newblock Image quality assessment: {U}nifying structure and texture similarity.
\newblock \emph{IEEE Transactions on Pattern Analysis and Machine Intelligence}, 44\penalty0 (5):\penalty0 2567--2581, 2022.

\bibitem[Fang et~al.(2017)Fang, Zhu, Ma, and Wang]{fang2017perceptual}
Yuming Fang, Hanwei Zhu, Kede Ma, and Zhou Wang.
\newblock Perceptual quality assessment of {HDR} deghosting algorithms.
\newblock In \emph{IEEE International Conference on Image Processing}, pages 3165--3169, 2017.

\bibitem[Fang et~al.(2020)Fang, Zhu, Zeng, Ma, and Wang]{fang2020perceptual}
Yuming Fang, Hanwei Zhu, Yan Zeng, Kede Ma, and Zhou Wang.
\newblock Perceptual quality assessment of smartphone photography.
\newblock In \emph{IEEE Conference on Computer Vision and Pattern Recognition}, pages 3677--3686, 2020.

\bibitem[Fang et~al.(2021)Fang, Zeng, Jiang, Zhu, and Yan]{fang2021superpixel}
Yuming Fang, Yan Zeng, Wenhui Jiang, Hanwei Zhu, and Jiebin Yan.
\newblock Superpixel-based quality assessment of multi-exposure image fusion for both static and dynamic scenes.
\newblock \emph{IEEE Transactions on Image Processing}, 30:\penalty0 2526--2537, 2021.

\bibitem[Huang et~al.(2025)Huang, Min, Ma, Liu, Zhou, Zhai, and et~al.]{diqa2025iccvw}
Fan Huang, Xiongkuo Min, Zhichao Ma, Xiaohong Liu, Chris~Wei Zhou, Guangtao Zhai, and et al.
\newblock Vquala 2025 document image quality assessment challenge.
\newblock In \emph{Proceedings of the IEEE/CVF Conference on Computer Vision (ICCV) Workshops}, pages 1--8, 2025.

\bibitem[Hui et~al.()Hui, Zhu, Yan, Liu, Jiang, and Zhao]{hui2024s}
Chen Hui, Haiqi Zhu, Shuya Yan, Shaohui Liu, Feng Jiang, and Debin Zhao.
\newblock S$^2$-csnet: Scale-aware scalable sampling network for image compressive sensing.
\newblock In \emph{ACM Multimedia}, pages 1--10.

\bibitem[Hui et~al.(2023)Hui, Zhang, Cui, Liu, Jiang, and Zhao]{hui2023rate}
Chen Hui, Shengping Zhang, Wenxue Cui, Shaohui Liu, Feng Jiang, and Debin Zhao.
\newblock Rate-adaptive neural network for image compressive sensing.
\newblock \emph{IEEE Transactions on Multimedia}, 26:\penalty0 2515--2530, 2023.

\bibitem[Hurst et~al.(2024)Hurst, Lerer, Goucher, Perelman, Ramesh, Clark, Ostrow, Welihinda, Hayes, Radford, et~al.]{hurst2024gpt}
Aaron Hurst, Adam Lerer, Adam~P Goucher, Adam Perelman, Aditya Ramesh, Aidan Clark, AJ Ostrow, Akila Welihinda, Alan Hayes, Alec Radford, et~al.
\newblock {GPT-4o} system card.
\newblock \emph{arXiv preprint arXiv:2410.21276}, 2024.

\bibitem[Huynh-Thu and Ghanbari(2008)]{psnr}
Quan Huynh-Thu and Mohammed Ghanbari.
\newblock Scope of validity of psnr in image/video quality assessment.
\newblock \emph{Electronics letters}, 44\penalty0 (13):\penalty0 800--801, 2008.

\bibitem[Kang et~al.(2014)Kang, Ye, Li, and Doermann]{kang2014convolutional}
Le Kang, Peng Ye, Yi Li, and David Doermann.
\newblock Convolutional neural networks for no-reference image quality assessment.
\newblock In \emph{Proceedings of the IEEE conference on computer vision and pattern recognition}, pages 1733--1740, 2014.

\bibitem[Ke et~al.(2021)Ke, Wang, Wang, Milanfar, and Yang]{ke2021musiq}
Junjie Ke, Qifei Wang, Yilin Wang, Peyman Milanfar, and Feng Yang.
\newblock {MUSIQ}: Multi-scale image quality transformer.
\newblock In \emph{IEEE International Conference on Computer Vision}, pages 5148--5157, 2021.

\bibitem[Kong et~al.(2022)Kong, Chen, Li, Wang, Rocha, and Kwong]{kong2022detect}
Chenqi Kong, Baoliang Chen, Haoliang Li, Shiqi Wang, Anderson Rocha, and Sam Kwong.
\newblock Detect and locate: Exposing face manipulation by semantic-and noise-level telltales.
\newblock \emph{IEEE Transactions on Information Forensics and Security}, 17:\penalty0 1741--1756, 2022.

\bibitem[Kong et~al.(2024)Kong, Luo, Bao, Yu, Li, Zheng, Wang, and Kot]{kong2024moe}
Chenqi Kong, Anwei Luo, Peijun Bao, Yi Yu, Haoliang Li, Zengwei Zheng, Shiqi Wang, and Alex~C Kot.
\newblock Moe-ffd: Mixture of experts for generalized and parameter-efficient face forgery detection.
\newblock \emph{arXiv preprint arXiv:2404.08452}, 2024.

\bibitem[Kong et~al.(2025)Kong, Luo, Wang, Li, Rocha, and Kot]{kong2025pixel}
Chenqi Kong, Anwei Luo, Shiqi Wang, Haoliang Li, Anderson Rocha, and Alex~C Kot.
\newblock Pixel-inconsistency modeling for image manipulation localization.
\newblock \emph{IEEE Transactions on Pattern Analysis and Machine Intelligence}, 2025.

\bibitem[Li et~al.(2024{\natexlab{a}})Li, Zhang, Guo, Zhang, Li, Zhang, Zhang, Zhang, Li, Liu, et~al.]{li2024llava}
Bo Li, Yuanhan Zhang, Dong Guo, Renrui Zhang, Feng Li, Hao Zhang, Kaichen Zhang, Peiyuan Zhang, Yanwei Li, Ziwei Liu, et~al.
\newblock {LLaVA-Onevision: E}asy visual task transfer.
\newblock \emph{arXiv preprint arXiv:2408.03326}, 2024{\natexlab{a}}.

\bibitem[Li et~al.(2025{\natexlab{a}})Li, Ma, Hua, Li, Wang, Zhou, Guan, Li, Yu, Lu, Liao, Ye, Chen, Sun, Cao, Cao, Zhang, Wen, Zhang, Chen, Lu, Min, Zhai, Xiao, Zhang, Su, Cheng, Liu, Xu, Chen, Hao, Zeng, Wu, Wang, Yu, Hu, Wang, Liu, Tong, Song, He, Wu, and Lyu]{li2025evqa}
Dasong Li, Sizhuo Ma, Hang Hua, Wenjie Li, Jian Wang, Chris~Wei Zhou, Fengbin Guan, Xin Li, Zihao Yu, Yiting Lu, Ru-Ling Liao, Yan Ye, Zhibo Chen, Wei Sun, Linhan Cao, Yuqin Cao, Weixia Zhang, Wen Wen, Kaiwei Zhang, Zijian Chen, Fangfang Lu, Xiongkuo Min, Guangtao Zhai, Erjia Xiao, Lingfeng Zhang, Zhenjie Su, Hao Cheng, Yu Liu, Renjing Xu, Long Chen, Xiaoshuai Hao, Zhenpeng Zeng, Jianqin Wu, Xuxu Wang, Qian Yu, Bo Hu, Weiwei Wang, Pinxin Liu, Yunlong Tong, Luchuan Song, Jinxi He, Jiaru Wu, and Hanjia Lyu.
\newblock Vquala 2025 challenge on engagement prediction for short videos: Methods and results.
\newblock In \emph{Proceedings of the IEEE/CVF Conference on Computer Vision (ICCV) Workshops}, pages 1--13, 2025{\natexlab{a}}.

\bibitem[Li et~al.(2025{\natexlab{b}})Li, Zhang, Zhao, Zhang, Li, Zhang, and Zhang]{qinsight}
Weiqi Li, Xuanyu Zhang, Shijie Zhao, Yabin Zhang, Junlin Li, Li Zhang, and Jian Zhang.
\newblock Q-insight: Understanding image quality via visual reinforcement learning.
\newblock \emph{arXiv preprint arXiv:2503.22679}, 2025{\natexlab{b}}.

\bibitem[Li et~al.(2024{\natexlab{b}})Li, Huang, Hu, Zhang, Cao, and Ji]{li2024boosting}
Xudong Li, Zihao Huang, Runze Hu, Yan Zhang, Liujuan Cao, and Rongrong Ji.
\newblock Boosting clip adaptation for image quality assessment via meta-prompt learning and gradient regularization.
\newblock \emph{arXiv preprint arXiv:2409.05381}, 2024{\natexlab{b}}.

\bibitem[Li et~al.(2025{\natexlab{c}})Li, Li, Zhou, Xing, Amirpour, Hao, Yue, Zhao, Liu, Yang, Tu, and et~al.]{isrgcq2025iccvw}
Yixiao Li, Xin Li, Chris~Wei Zhou, Shuo Xing, Hadi Amirpour, Xiaoshuai Hao, Guanghui Yue, Baoquan Zhao, Weide Liu, Xiaoyuan Yang, Zhengzhong Tu, and et al.
\newblock Vquala 2025 challenge on image super-resolution generated content quality assessment: Methods and results.
\newblock In \emph{Proceedings of the IEEE/CVF Conference on Computer Vision (ICCV) Workshops}, pages 1--10, 2025{\natexlab{c}}.

\bibitem[Liao et~al.(2022)Liao, Chen, Zhu, Wang, Zhou, and Kwong]{deepwsd}
Xingran Liao, Baoliang Chen, Hanwei Zhu, Shiqi Wang, Mingliang Zhou, and Sam Kwong.
\newblock {DeepWSD: P}rojecting degradations in perceptual space to {Wasserstein} distance in deep feature space.
\newblock In \emph{ACM International Conference on Multimedia}, pages 970--978, 2022.

\bibitem[Lin and Kuo(2011)]{lin2011perceptual}
Weisi Lin and C-C~Jay Kuo.
\newblock Perceptual visual quality metrics: A survey.
\newblock \emph{Journal of Visual Communication and Image Representation}, 22\penalty0 (4):\penalty0 297--312, 2011.

\bibitem[Liu et~al.(2017)Liu, Van De~Weijer, and Bagdanov]{liu2017rankiqa}
Xialei Liu, Joost Van De~Weijer, and Andrew~D Bagdanov.
\newblock Rankiqa: Learning from rankings for no-reference image quality assessment.
\newblock In \emph{Proceedings of the IEEE international conference on computer vision}, pages 1040--1049, 2017.

\bibitem[Liu et~al.(2023)Liu, Yan, Wan, Fang, and Wang]{liu2023quality}
Xuelin Liu, Jiebin Yan, Zheng Wan, Yuming Fang, and Zhou Wang.
\newblock A quality-of-experience database for adaptive omnidirectional video streaming.
\newblock \emph{IEEE Journal of Selected Topics in Signal Processing}, 17\penalty0 (5):\penalty0 949--963, 2023.

\bibitem[Ma et~al.(2025)Ma, Chen, Gao, Wang, Zhou, Sun, Zhang, Cao, Jia, Zhu, Zhu, Min, Zhai, Chen, Xiao, Zeng, Wu, Lou, Tan, Song, Xu, Hamidi, Amirpour, Bai, Du, Jiang, Lu, Cui, Gan, Li, Jiang, Li, Wang, Yuan, Li, Chen, Deng, Deng, Chen, Yao, Zheng, Zhang, Fu, Joshi, Agarwal, Immidisetti, Mopidevi, Shukla, Yang, Zhang, Pan, Deng, Ouyang, Yang, Luo, Shi, Lai, Ruan, and Yue]{ma2025fiqa}
Sizhuo Ma, Wei-Ting Chen, Qiang Gao, Jian Wang, Chris~Wei Zhou, Wei Sun, Weixia Zhang, Linhan Cao, Jun Jia, Xiangyang Zhu, Dandan Zhu, Xiongkuo Min, Guangtao Zhai, Baoying Chen, Xiongwei Xiao, Jishen Zeng, Wei Wu, Tiexuan Lou, Yuchen Tan, Chunyi Song, Zhiwei Xu, MohammadAli Hamidi, Hadi Amirpour, Mingyin Bai, Jiawang Du, Zhenyu Jiang, Zilong Lu, Ziguan Cui, Zongliang Gan, Xinpeng Li, Shiqi Jiang, Chenhui Li, Changbo Wang, Weijun Yuan, Zhan Li, Yihang Chen, Yifan Deng, Ruting Deng, Zhanglu Chen, Boyang Yao, Shuling Zheng, Feng Zhang, Zhiheng Fu, Abhishek Joshi, Aman Agarwal, Rakhil Immidisetti, Ajay~Narasimha Mopidevi, Vishwajeet Shukla, Hao Yang, Ruikun Zhang, Liyuan Pan, Kaixin Deng, Hang Ouyang, Fan Yang, Zhizun Luo, Zhuohang Shi, Songning Lai, Weilin Ruan, and Yutao Yue.
\newblock Vquala 2025 challenge on face image quality assessment: Methods and results.
\newblock In \emph{Proceedings of the IEEE/CVF Conference on Computer Vision (ICCV) Workshops}, pages 1--10, 2025.

\bibitem[Madhusudana et~al.(2022)Madhusudana, Birkbeck, Wang, Adsumilli, and Bovik]{madhusudana2022image}
Pavan~C Madhusudana, Neil Birkbeck, Yilin Wang, Balu Adsumilli, and Alan~C Bovik.
\newblock Image quality assessment using contrastive learning.
\newblock \emph{IEEE Transactions on Image Processing}, 31:\penalty0 4149--4161, 2022.

\bibitem[Prashnani et~al.(2018)Prashnani, Cai, Mostofi, and Sen]{prashnani2018pieapp}
Ekta Prashnani, Hong Cai, Yasamin Mostofi, and Pradeep Sen.
\newblock {PieAPP: P}erceptual image-error assessment through pairwise preference.
\newblock In \emph{IEEE Conference on Computer Vision and Pattern Recognition}, pages 1808--1817, 2018.

\bibitem[Sheikh and Bovik(2006)]{sheikh2006image}
Hamid~R Sheikh and Alan~C. Bovik.
\newblock Image information and visual quality.
\newblock \emph{IEEE Transactions on Image Processing}, 15\penalty0 (2):\penalty0 430--444, 2006.

\bibitem[Su et~al.(2020)Su, Yan, Zhu, Zhang, Ge, Sun, and Zhang]{hyperiqa}
Shaolin Su, Qingsen Yan, Yu Zhu, Cheng Zhang, Xin Ge, Jinqiu Sun, and Yanning Zhang.
\newblock Blindly assess image quality in the wild guided by a self-adaptive hyper network.
\newblock In \emph{IEEE Conference on Computer Vision and Pattern Recognition}, pages 3664--3673, 2020.

\bibitem[Sui et~al.(2023)Sui, Zhu, Liu, Fang, Wang, and Wang]{sui2023perceptual}
Xiangjie Sui, Hanwei Zhu, Xuelin Liu, Yuming Fang, Shiqi Wang, and Zhou Wang.
\newblock Perceptual quality assessment of 360 images based on generative scanpath representation.
\newblock \emph{arXiv preprint arXiv:2309.03472}, 2023.

\bibitem[Tian et~al.(2025)Tian, Li, Chen, Zhu, Wang, and Kwong]{tian2025ai}
Yu Tian, Yixuan Li, Baoliang Chen, Hanwei Zhu, Shiqi Wang, and Sam Kwong.
\newblock {AI}-generated image quality assessment in visual communication.
\newblock In \emph{AAAI Conference on Artificial Intelligence}, pages 7392--7400, 2025.

\bibitem[Wang et~al.(2022)Wang, Fan, Hou, Xu, Li, Lu, and Fu]{wang2022mstriq}
Jing Wang, Haotian Fan, Xiaoxia Hou, Yitian Xu, Tao Li, Xuechao Lu, and Lean Fu.
\newblock Mstriq: No reference image quality assessment based on swin transformer with multi-stage fusion.
\newblock In \emph{Proceedings of the IEEE/CVF Conference on Computer Vision and Pattern Recognition}, pages 1269--1278, 2022.

\bibitem[Wang et~al.(2004{\natexlab{a}})Wang, Bovik, Sheikh, and Simoncelli]{ssim}
Zhou Wang, Alan~C Bovik, Hamid~R Sheikh, and Eero~P Simoncelli.
\newblock Image quality assessment: from error visibility to structural similarity.
\newblock \emph{IEEE Transactions on Image Processing}, 13\penalty0 (4):\penalty0 600--612, 2004{\natexlab{a}}.

\bibitem[Wang et~al.(2004{\natexlab{b}})Wang, Bovik, Sheikh, and Simoncelli]{wang2004image}
Zhou Wang, Alan~C. Bovik, Hamid~R Sheikh, and Eero~P. Simoncelli.
\newblock Image quality assessment: From error visibility to structural similarity.
\newblock \emph{IEEE Transactions on Image Processing}, 13\penalty0 (4):\penalty0 600--612, 2004{\natexlab{b}}.

\bibitem[Wu et~al.(2024{\natexlab{a}})Wu, Zhang, Zhang, Chen, Liao, Wang, Xu, Li, Hou, Zhai, Xue, Sun, Yan, and Lin]{Wu_2024_CVPR}
Haoning Wu, Zicheng Zhang, Erli Zhang, Chaofeng Chen, Liang Liao, Annan Wang, Kaixin Xu, Chunyi Li, Jingwen Hou, Guangtao Zhai, Geng Xue, Wenxiu Sun, Qiong Yan, and Weisi Lin.
\newblock {Q-Instruct: I}mproving low-level visual abilities for multi-modality foundation models.
\newblock In \emph{IEEE Conference on Computer Vision and Pattern Recognition}, pages 25490--25500, 2024{\natexlab{a}}.

\bibitem[Wu et~al.(2024{\natexlab{b}})Wu, Zhang, Zhang, Chen, Liao, Li, Gao, Wang, Zhang, Sun, Yan, Min, Zhai, and Lin]{qalign}
Haoning Wu, Zicheng Zhang, Weixia Zhang, Chaofeng Chen, Liang Liao, Chunyi Li, Yixuan Gao, Annan Wang, Erli Zhang, Wenxiu Sun, Qiong Yan, Xiongkuo Min, Guangtao Zhai, and Weisi Lin.
\newblock {Q-Align: T}eaching {LMM}s for visual scoring via discrete text-defined levels.
\newblock In \emph{International Conference on Machine Learning}, pages 54015--54029, 2024{\natexlab{b}}.

\bibitem[Wu et~al.(2024{\natexlab{c}})Wu, Zhu, Zhang, Zhang, Chen, Liao, Li, Wang, Sun, Yan, et~al.]{wu2024towards}
Haoning Wu, Hanwei Zhu, Zicheng Zhang, Erli Zhang, Chaofeng Chen, Liang Liao, Chunyi Li, Annan Wang, Wenxiu Sun, Qiong Yan, et~al.
\newblock Towards open-ended visual quality comparison.
\newblock In \emph{European Conference on Computer Vision}, pages 360--377, 2024{\natexlab{c}}.

\bibitem[Wu et~al.(2024{\natexlab{d}})Wu, Ma, Liang, Yang, and Zhang]{wu2024comprehensive}
Tianhe Wu, Kede Ma, Jie Liang, Yujiu Yang, and Lei Zhang.
\newblock A comprehensive study of multimodal large language models for image quality assessment.
\newblock In \emph{European Conference on Computer Vision}, pages 143--160. Springer, 2024{\natexlab{d}}.

\bibitem[Wu et~al.(2025)Wu, Zou, Liang, Zhang, and Ma]{wu2025visualquality}
Tianhe Wu, Jian Zou, Jie Liang, Lei Zhang, and Kede Ma.
\newblock Visualquality-r1: Reasoning-induced image quality assessment via reinforcement learning to rank.
\newblock \emph{arXiv preprint arXiv:2505.14460}, 2025.

\bibitem[Yan et~al.(2022)Yan, Li, Fang, Che, Xia, and Liu]{yan2022subjective}
Jiebin Yan, Jing Li, Yuming Fang, Zhaohui Che, Xue Xia, and Yang Liu.
\newblock Subjective and objective quality of experience of free viewpoint videos.
\newblock \emph{IEEE Transactions on Image Processing}, 31:\penalty0 3896--3907, 2022.

\bibitem[Ying et~al.(2020)Ying, Niu, Gupta, Mahajan, Ghadiyaram, and Bovik]{ying2020patches}
Zhenqiang Ying, Haoran Niu, Praful Gupta, Dhruv Mahajan, Deepti Ghadiyaram, and Alan Bovik.
\newblock From patches to pictures (paq-2-piq): Mapping the perceptual space of picture quality.
\newblock In \emph{Proceedings of the IEEE/CVF conference on computer vision and pattern recognition}, pages 3575--3585, 2020.

\bibitem[You et~al.(2024)You, Li, Gu, Yin, Xue, and Dong]{you2024depicting}
Zhiyuan You, Zheyuan Li, Jinjin Gu, Zhenfei Yin, Tianfan Xue, and Chao Dong.
\newblock Depicting beyond scores: {A}dvancing image quality assessment through multi-modal language models.
\newblock In \emph{European Conference on Computer Vision}, pages 259--276, 2024.

\bibitem[You et~al.(2025)You, Cai, Gu, Xue, and Dong]{you2025teaching}
Zhiyuan You, Xin Cai, Jinjin Gu, Tianfan Xue, and Chao Dong.
\newblock Teaching large language models to regress accurate image quality scores using score distribution.
\newblock In \emph{IEEE Conference on Computer Vision and Pattern Recognition}, 2025.

\bibitem[Zhang et~al.(2018)Zhang, Isola, Efros, Shechtman, and Wang]{LPIPS18}
Richard Zhang, Phillip Isola, Alexei~A Efros, Eli Shechtman, and Oliver Wang.
\newblock The unreasonable effectiveness of deep features as a perceptual metric.
\newblock In \emph{IEEE Conference on Computer Vision and Pattern Recognition}, pages 586--595, 2018.

\bibitem[Zhang et~al.(2020)Zhang, Ma, Yan, Deng, and Wang]{dbcnn}
Weixia Zhang, Kede Ma, Jia Yan, Dexiang Deng, and Zhou Wang.
\newblock Blind image quality assessment using a deep bilinear convolutional neural network.
\newblock \emph{IEEE Transactions on Circuits and Systems for Video Technology}, 30\penalty0 (1):\penalty0 36--47, 2020.

\bibitem[Zhang et~al.(2021)Zhang, Ma, Zhai, and Yang]{unique}
Weixia Zhang, Kede Ma, Guangtao Zhai, and Xiaokang Yang.
\newblock Uncertainty-aware blind image quality assessment in the laboratory and wild.
\newblock \emph{IEEE Transactions on Image Processing}, 30:\penalty0 3474--3486, 2021.

\bibitem[Zhang et~al.(2023)Zhang, Zhai, Wei, Yang, and Ma]{liqe}
Weixia Zhang, Guangtao Zhai, Ying Wei, Xiaokang Yang, and Kede Ma.
\newblock Blind image quality assessment via vision-language correspondence: A multitask learning perspective.
\newblock In \emph{IEEE Conference on Computer Vision and Pattern Recognition}, pages 14071--14081, 2023.

\bibitem[Zhao et~al.(2018)Zhao, Liu, Kong, Zhao, Guo, Liu, Ding, Ding, Tan, and Li]{zhao2018faster}
Weisong Zhao, Jian Liu, Chenqi Kong, Yixuan Zhao, Changliang Guo, Chenguang Liu, Xiangyan Ding, Xumin Ding, Jiubin Tan, and Haoyu Li.
\newblock Faster super-resolution imaging with auto-correlation two-step deconvolution.
\newblock \emph{arXiv preprint arXiv:1809.07410}, 2018.

\bibitem[Zhu et~al.(2022{\natexlab{a}})Zhu, Chen, Zhu, and Wang]{zhu2022learning}
Hanwei Zhu, Baoliang Chen, Lingyu Zhu, and Shiqi Wang.
\newblock Learning spatiotemporal interactions for user-generated video quality assessment.
\newblock \emph{IEEE Transactions on Circuits and Systems for Video Technology}, 33\penalty0 (3):\penalty0 1031--1042, 2022{\natexlab{a}}.

\bibitem[Zhu et~al.(2022{\natexlab{b}})Zhu, Chen, Zhu, Wang, and Lin]{zhu2022deepdc}
Hanwei Zhu, Baoliang Chen, Lingyu Zhu, Shiqi Wang, and Weisi Lin.
\newblock {DeepDC: D}eep distance correlation as a perceptual image quality evaluator.
\newblock \emph{arXiv preprint arXiv:2211.04927}, 2022{\natexlab{b}}.

\bibitem[Zhu et~al.(2024{\natexlab{a}})Zhu, Chen, Zhu, Chen, Song, and Wang]{zhu2024video}
Hanwei Zhu, Baoliang Chen, Lingyu Zhu, Peilin Chen, Linqi Song, and Shiqi Wang.
\newblock Video quality assessment for spatio-temporal resolution adaptive coding.
\newblock \emph{IEEE Transactions on Circuits and Systems for Video Technology}, 2024{\natexlab{a}}.

\bibitem[Zhu et~al.(2024{\natexlab{b}})Zhu, Sui, Chen, Liu, Chen, Fang, and Wang]{zhu2afc24}
Hanwei Zhu, Xiangjie Sui, Baoliang Chen, Xuelin Liu, Peilin Chen, Yuming Fang, and Shiqi Wang.
\newblock 2afc prompting of large multimodal models for image quality assessment.
\newblock \emph{IEEE Transactions on Circuits and Systems for Video Technology}, 34\penalty0 (12):\penalty0 12873--12878, 2024{\natexlab{b}}.

\bibitem[Zhu et~al.(2024{\natexlab{c}})Zhu, Wu, Li, Zhang, Chen, Zhu, Fang, Zhai, Lin, and Wang]{zhu2024adaptive}
Hanwei Zhu, Haoning Wu, Yixuan Li, Zicheng Zhang, Baoliang Chen, Lingyu Zhu, Yuming Fang, Guangtao Zhai, Weisi Lin, and Shiqi Wang.
\newblock Adaptive image quality assessment via teaching large multimodal model to compare.
\newblock In \emph{Advances in Neural Information Processing Systems}, pages 32611--32629, 2024{\natexlab{c}}.

\bibitem[Zhu et~al.(2024{\natexlab{d}})Zhu, Yang, Chen, Zhu, Meng, and Wang]{zhu2024temporally}
Lingyu Zhu, Wenhan Yang, Baoliang Chen, Hanwei Zhu, Xiandong Meng, and Shiqi Wang.
\newblock Temporally consistent enhancement of low-light videos via spatial-temporal compatible learning.
\newblock \emph{International Journal of Computer Vision}, 132\penalty0 (10):\penalty0 4703--4723, 2024{\natexlab{d}}.

\bibitem[Zhu et~al.(2024{\natexlab{e}})Zhu, Yang, Chen, Zhu, Ni, Mao, and Wang]{zhu2024unrolled}
Lingyu Zhu, Wenhan Yang, Baoliang Chen, Hanwei Zhu, Zhangkai Ni, Qi Mao, and Shiqi Wang.
\newblock Unrolled decomposed unpaired learning for controllable low-light video enhancement.
\newblock In \emph{European Conference on Computer Vision}, pages 329--347. Springer, 2024{\natexlab{e}}.

\end{thebibliography}
}

\appendix

\subsection*{Organizers}
\label{sec:organizers}
\noindent\textit{\textbf{Title:}} VQualA 2025 Challenge on Visual Quality Comparison for Large Multimodal Models: Methods and Results

\noindent\textit{\textbf{Members:}} \\
Hanwei Zhu\textsuperscript{1} (\textcolor{magenta}{hanwei.zhu@ntu.edu.sg}), \\
Haoning Wu\textsuperscript{2} (\textcolor{magenta}{haoning001@e.ntu.edu.sg}),\\
Zicheng Zhang\textsuperscript{3} (\textcolor{magenta}{zzc1998@sjtu.edu.cn}), \\
Lingyu Zhu\textsuperscript{4} (\textcolor{magenta}{lingyzhu-c@my.cityu.edu.hk}), \\
Yixuan Li \textsuperscript{4} (\textcolor{magenta}{yixuanli423@gmail.com}), \\ 
Peilin Chen\textsuperscript{4} (\textcolor{magenta}{plchen3@cityu.edu.hk}), \\ 
Shiqi Wang\textsuperscript{4} (\textcolor{magenta}{shiqwang@cityu.edu.hk}), \\
Chris Wei Zhou\textsuperscript{5} (\textcolor{magenta}{zhouw26@cardiff.ac.uk}), \\

\noindent\textit{\textbf{Affiliations:}}

\noindent\textsuperscript{1} Nanyang Technological
University

\noindent\textsuperscript{2} Moonshot AI

\noindent\textsuperscript{3} Shanghai Jiao Tong University

\noindent\textsuperscript{4} City University of Hong Kong

\noindent\textsuperscript{5} Cardiff University

\subsection*{ECNU-SJTU VQA Team}

\noindent\textit{\textbf{Title:}}  Fine-Grained Visual Quality Comparison via Ensemble Voting of Large Multimodal Models

\noindent\textit{\textbf{Members:}} Linhan Cao\textsuperscript{1} (\textcolor{magenta}{caolinhan@sjtu.edu.cn}), Wei Sun\textsuperscript{2}, Xiangyang Zhu\textsuperscript{3}, Weixia Zhang\textsuperscript{1}, Yucheng Zhu\textsuperscript{1}, Jing Liu\textsuperscript{4}, Dandan Zhu\textsuperscript{2}, Guangtao Zhai\textsuperscript{1}, Xiongkuo Min\textsuperscript{1}

\noindent\textit{\textbf{Affiliations:}}

\noindent\textsuperscript{1} Shanghai Jiao Tong University

\noindent\textsuperscript{2} East China Normal University

\noindent\textsuperscript{3} Shanghai Artificial Intelligence Laboratory    

\noindent\textsuperscript{4} Tianjin University

\subsection*{Digital Ocean}

\noindent\textit{\textbf{Title:}} Joint Optimization of Visual Quality Comparison through Multi-modal Large Language Model

\noindent\textit{\textbf{Members:}} Zhichao Zhang\textsuperscript{1} (\textcolor{magenta}{liquortect@sjtu.edu.cn}), Xinyue Li\textsuperscript{1}, Shubo Xu\textsuperscript{2}

\noindent\textit{\textbf{Affiliations:}}

\noindent\textsuperscript{1}Shanghai Jiao Tong University

\noindent\textsuperscript{2}Baidu Inc.

\subsection*{ActionLab}
\noindent\textit{\textbf{Title:}} Two-Stage Fine-Tuning of QwenVL2.5-7B for Open-Ended Visual Quality Comparison

\noindent\textit{\textbf{Members:}} Anh Dao~(\textcolor{magenta}{anhdao@msu.edu}), Yifan Li

\noindent\textit{\textbf{Affiliations:}}

\noindent Michigan State University

\subsection*{XiaomiMM}

\noindent\textit{\textbf{Title:}} Enhancing Multi-Image Reasoning Abilities of Large Multi-modal Models via Auxiliary Visual Scoring

\noindent\textit{\textbf{Members:}} Hongyuan Yu\textsuperscript{1} (\textcolor{magenta}{yuhyuan1995@gmail.com}), Jiaojiao Yi\textsuperscript{1}, Yiding Tian\textsuperscript{2}, Yupeng Wu\textsuperscript{3}, Feiran Sun,\textsuperscript{1}, Yuhui Wu\textsuperscript{1}

\noindent\textit{\textbf{Affiliations:}}

\noindent\textsuperscript{1}Multimedia Department, Xiaomi Inc.

\noindent\textsuperscript{2}Xiamen University

\noindent\textsuperscript{3}Institute of Automation, Chinese Academy of Sciences

\subsection*{Labubu}

\noindent\textit{\textbf{Title:}} Exploring MLLM in Fine-Grained Visual Quality Comparison with Quality Token

\noindent\textit{\textbf{Members:}} Lijuan Jiao (\textcolor{magenta}{lijuan.jiao@samsung.com}), Song Jiang 

\noindent\textit{\textbf{Affiliations:}}

\noindent Samsung R\&D Institute China Xi’an

\end{document}